\documentclass[letterpaper]{article}
\usepackage{aaai2027}
\nocopyright  
\usepackage{times}
\usepackage{helvet}
\usepackage{courier}
\usepackage[hyphens]{url}
\usepackage{graphicx}
\usepackage{natbib}
\usepackage{caption}
\usepackage{booktabs}
\usepackage{amsmath}
\usepackage{amssymb}
\usepackage{algorithm}
\usepackage{algorithmic}
\graphicspath{{./}}

\title{PhenoStitch: Training-Free Panoptic Crop Mapping\\from Satellite Image Time Series}
\author{
    Xuechen Li
}
\affiliations{
    University of Minnesota\\
    Minneapolis, MN, USA\\
    li003487@umn.edu
}

\begin{document}
\maketitle

\begin{abstract}
Panoptic crop mapping requires both delineating individual agricultural parcels and assigning a
crop type to each parcel from satellite image time series. Existing approaches typically rely on
dense parcel-level annotations and task-specific model training, which limits their applicability
to new regions and growing seasons. We introduce \emph{PhenoStitch}, a panoptic crop-mapping
pipeline that requires no task-specific gradient-based training. A frozen Segment Anything model
first over-segments each patch into class-agnostic regions. For each region, optical NDVI and
Sentinel-1 backscatter series are summarized by an analytic double-harmonic phenological
signature. Adjacent regions are then merged into parcels by minimizing a Potts graph energy, and
each parcel is classified by nearest-prototype matching using only $k$ labeled parcels per class.
A final topology-closure step produces the panoptic map. Under a matched budget of $k{=}20$
parcels per class, corresponding to less than $1\%$ of the available labels, PhenoStitch achieves
$20.0$ crop mIoU, $76.2$ segmentation quality, and $6.2$ panoptic quality on PASTIS-R under
$5$-fold, $3$-seed evaluation. It outperforms the evaluated frozen foundation-model, few-shot,
and matched-budget supervised baselines under the same protocol. A consistent ranking is observed
on ZueriCrop. Ablations show that radar observations contribute the largest gain, with the
graph-energy merge and a compact phenological signature helping further. The results demonstrate the effectiveness of combining label-free parcel
delineation with few-shot phenological recognition for panoptic crop mapping under limited
supervision.
\end{abstract}

\section{Introduction}

Turning a stack of satellite images into a crop map means answering two questions at once:
\emph{where} is each field, and \emph{what} grows in it. In panoptic terms, the first
is instance segmentation of parcels and the second is semantic labeling of crop type, and a
useful map needs both. The dominant approach trains deep spatio-temporal networks end-to-end
on densely annotated satellite image time series (SITS)~\citep{garnot2021panoptic,
tarasiou2023tsvit}. These models are strong, but only where dense labels exist. Such labels
are missing exactly where mapping matters most: new countries, new years, minor crops, and
smallholder landscapes~\citep{rustowicz2019}. Parcel-level ground truth must be re-collected
whenever the region or season changes. A method that collapses without thousands of labeled
parcels therefore cannot map the places that need it most.

\begin{figure}[t]
\centering
\includegraphics[width=\columnwidth]{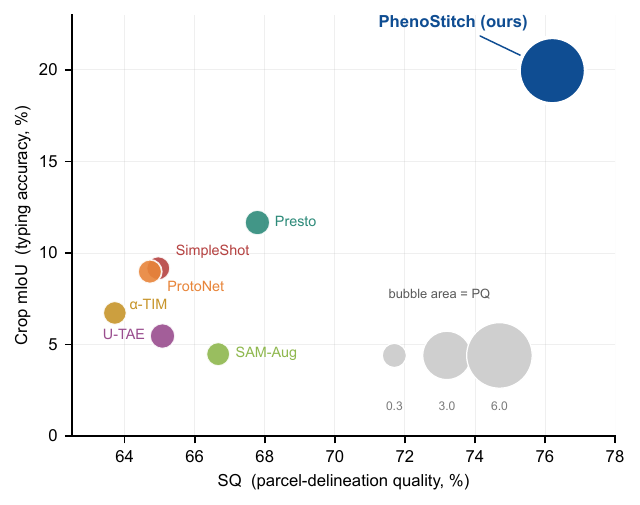}
\caption{\textbf{Comparison under a matched label-scarce protocol on PASTIS-R.}
Each bubble represents one method at $k{=}20$ using $5$-fold$\times3$-seed
cross-validation. The horizontal axis reports segmentation quality (SQ), the vertical axis
reports crop mIoU, and bubble area represents panoptic quality (PQ). For semantic baselines,
parcel instances are obtained using the same class-wise connected-component post-processing.
PhenoStitch achieves the highest SQ, crop mIoU, and PQ among the compared methods. TSViT is
omitted for visual clarity because its SQ is approximately $4$.}
\label{fig:hero}
\end{figure}

This motivates the \emph{label-scarce} regime: at most a handful of labeled parcels per crop
class, with no chance to train a large model. Existing options degrade sharply here. Deep
SITS networks trained on the same tiny budget overfit and lose most of their accuracy.
Few-shot classifiers and frozen remote-sensing foundation models keep more accuracy, but they
classify pixels, not parcels. Their maps can be turned into instances by connected components,
the usual stand-in for a learned parcel detector. Even then, the resulting regions rarely
match true parcels. The panoptic quality (PQ) of every baseline we test therefore stays near
zero, whatever its semantic accuracy.

Our starting point is that the two hard parts of crop mapping are each solvable without any
training. Boundaries are a generic vision primitive: a frozen promptable segmenter splits a
field patch into coherent regions from appearance alone, with no crop labels. Crop identity,
in turn, is written into \emph{phenology}, the seasonal trajectory of growth, senescence, and
moisture that separates wheat from maize~\citep{turkoglu2021zuericrop,veloso2017}. This
trajectory is directly observable: an optical vegetation index and a radar backscatter series,
fit by a low-order harmonic model, give each region a compact, calibration-free seasonal
signature. Neither step needs gradient training, and both are exactly what a panoptic map
requires.

We turn this observation into \emph{PhenoStitch}, a fully training-free pipeline for panoptic
crop mapping (Figure~\ref{fig:pipeline}). A frozen Segment Anything model first over-segments
each patch. Every region is then described by an analytic double-harmonic phenological
signature of its NDVI and Sentinel-1 backscatter. Regions are merged into field parcels by
minimizing a Potts graph energy on these signatures. Each parcel is typed by nearest-prototype
matching to only $k$ labeled parcels per class, and a topology-closure step produces the
panoptic map. No component is trained by gradient descent. Our contributions are: (i) a training-free formulation of panoptic crop
mapping that decouples label-free delineation from few-shot typing; (ii) a phenological
signature and graph-energy merge that turn generic over-segmentation into agronomically
coherent parcels; and (iii) a rigorous label-scarce evaluation, using $5$-fold $\times\,3$-seed
cross-validation on PASTIS-R and a second dataset, ZueriCrop. This evaluation shows that
PhenoStitch is state-of-the-art on \emph{every} panoptic metric at $\sim\!1\%$ labels
(Figure~\ref{fig:hero}), while remaining explicit about where its accuracy is capped.

\section{Related Work}

\paragraph{Supervised crop mapping from SITS.}
Crop mapping from SITS is dominated by supervised temporal and spatio-temporal encoders:
recurrent encoders~\citep{russwurm2018recurrent}, temporal convolutions
(TempCNN)~\citep{pelletier2019tempcnn}, multi-source optical--radar
fusion~\citep{ienco2019combining}, raw-series self-attention~\citep{russwurm2020selfattention}
on datasets such as BreizhCrops~\citep{russwurm2020breizhcrops} and
Sen4AgriNet~\citep{sykas2021sen4agrinet}, pixel-set encoders with lightweight temporal
attention~\citep{garnot2020pse}, U-TAE~\citep{garnot2021panoptic}, and time-first vision
transformers (TSViT)~\citep{tarasiou2023tsvit}. For the \emph{panoptic} variant, U-TAE is
paired with a parcel detector (PaPs) on PASTIS~\citep{garnot2021panoptic}, and label
hierarchies are exploited on ZueriCrop~\citep{turkoglu2021zuericrop}. These methods set the
state of the art \emph{when} dense labels are available; on a matched few-label budget they
overfit, as our experiments confirm.

\paragraph{Label-efficient methods and panoptic evaluation.}
Two routes reduce label cost, but both stop at semantics. Self-supervised remote-sensing
foundation models pre-train on unlabeled imagery and transfer: seasonal contrast
(SeCo)~\citep{manas2021seco}, SatMAE~\citep{cong2022satmae}, Prithvi~\citep{jakubik2023prithvi},
Scale-MAE~\citep{reed2023scalemae}, the optical--radar model CROMA~\citep{fuller2023croma},
DOFA~\citep{xiong2024dofa}, and the pixel-timeseries model Presto~\citep{tseng2023presto}, with
cross-region temporal adaptation easing domain shift~\citep{nyborg2022timematch}. Orthogonally,
few-shot learners classify from a handful of exemplars: matching and prototypical
networks~\citep{vinyals2016matching,snell2017protonet}, gradient-based
meta-learning~\citep{finn2017maml}, strong nearest-neighbor
baselines~\citep{wang2019simpleshot,chen2019closerlook}, prototype-aligned few-shot
segmentation (PANet)~\citep{wang2019panet}, and transductive information maximization
($\alpha$-TIM)~\citep{veilleux2021alphatim}. Both are per-pixel semantic classifiers that yield
no parcel instances, and they are our strongest baselines. Panoptic segmentation instead scores
instances-with-types through $\mathrm{PQ}=\mathrm{SQ}\times\mathrm{RQ}$~\citep{kirillov2019panoptic},
with architectures such as Panoptic-DeepLab~\citep{cheng2020panopticdeeplab} and mask
transformers~\citep{cheng2021maskformer,cheng2022mask2former} built for label-rich natural
images. Converting a semantic map to instances by connected components leaves
$\mathrm{RQ}\!\approx\!0$ whenever those regions do not match true parcels, the regime every
semantic baseline falls into.

\paragraph{Object-based delineation and phenology.}
PhenoStitch instead assembles training-free ingredients that already exist separately.
Object-based image analysis has long segmented imagery into objects and then classified
them~\citep{blaschke2010obia}, and field-boundary delineation recovers parcel instances but not
crop type~\citep{waldner2021fields}. Promptable segmenters supply such objects from appearance
alone: Segment Anything (SAM)~\citep{kirillov2023sam} and higher-quality
variants~\citep{ke2023hqsam}, adapted to overhead imagery through large mask
datasets~\citep{wang2023samrs}, learned prompting~\citep{chen2024rsprompter}, or mask-augmented
training~\citep{samaug2026}, while open-vocabulary segmenters built on
CLIP~\citep{radford2021clip} add zero-shot semantics, from dense CLIP labels~\citep{zhou2022maskclip}
and LSeg~\citep{li2022lseg} to remote-sensing SegEarth-OV~\citep{li2025segearth}. None of these
assign crop type. Crop identity instead lives in phenology, which harmonic and change-detection
models read from vegetation time series: CCDC~\citep{zhu2014ccdc},
BFAST~\citep{verbesselt2010bfast}, and joint optical--radar crop dynamics~\citep{veloso2017}.
PhenoStitch is the first to combine a frozen SAM over-segmentation with an analytic phenological
signature and few-shot typing, so that parcels and their crop types fall out of a single
training-free panoptic pipeline.

\section{Method}

\begin{figure*}[t]
\centering
\includegraphics[width=\textwidth]{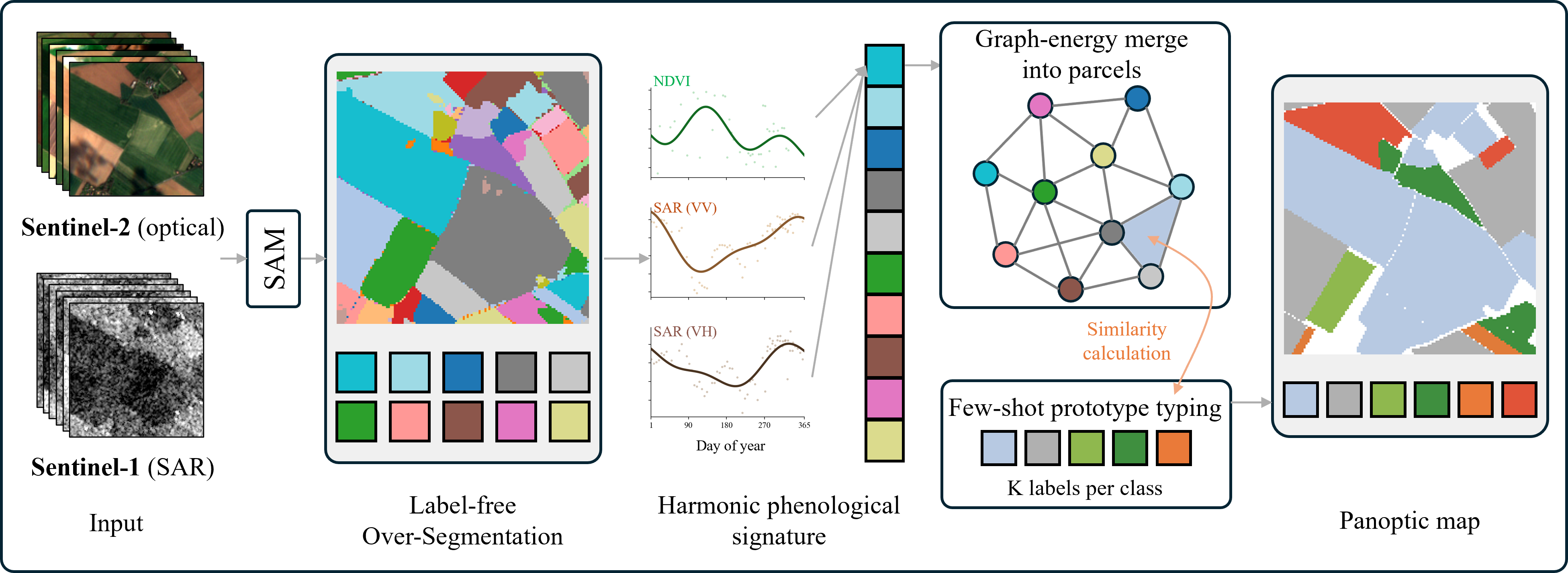}
\caption{\textbf{The PhenoStitch pipeline is training-free.} From a Sentinel-2 optical and
Sentinel-1 SAR image time series (\emph{Input}): a frozen SAM over-segments each patch into
candidate regions (\emph{label-free over-segmentation}); each region's NDVI and SAR (VV/VH) series
are summarized by an analytic double-harmonic signature (\emph{harmonic phenological signature});
adjacent regions with similar signatures are merged into parcels by a Potts graph energy
(\emph{graph-energy merge into parcels}); and each parcel is typed by nearest-prototype matching
to $k$ labels per class, the only labels used (\emph{few-shot prototype typing}),
yielding the \emph{panoptic} crop map of parcel instances and crop types. No network is
trained end-to-end.}
\label{fig:pipeline}
\end{figure*}

\paragraph{Task and notation.}
The input is a co-registered SITS $X=\{(I_t,\tau_t)\}_{t=1}^{T}$ over one patch, where $I_t$
stacks Sentinel-2 (optical) and Sentinel-1 (SAR) channels acquired at day-of-year $\tau_t$.
The output is a panoptic map assigning every pixel a parcel instance and a crop class
$c\in\{1,\dots,C\}$. The only supervision is a small support set of $k$ labeled parcels per
class ($k$-shot); delineation uses \emph{zero} labels. Figure~\ref{fig:pipeline} shows the
five stages.

\paragraph{Design rationale.}
The pipeline is organized around one observation: the two questions of panoptic crop mapping draw
on different information. \emph{Where} a field lies depends on appearance and geometry, which a
promptable segmenter captures with no crop labels. \emph{What} grows there depends on temporal
phenology, which a compact harmonic descriptor captures with no trained classifier. Keeping the
two legs separate is what makes the method training-free. Neither leg needs gradient supervision,
and each can be validated on its own, delineation by SQ and typing by mIoU. The split also shapes
the error profile reported later: delineation is stable because appearance is easy, whereas
typing is bounded by crops whose phenology genuinely overlaps.

\paragraph{(A) Label-free over-segmentation.}
A crop parcel is homogeneous in appearance, so its boundary is a generic signal rather than a
crop-specific one. Unlike classical unsupervised over-segmentation such as graph-based
segmentation~\citep{felzenszwalb2004} or SLIC superpixels~\citep{achanta2012slic}, a frozen
SAM produces boundaries that already respect object structure. We run a frozen
SAM~\citep{kirillov2023sam} on a temporal composite of the patch and obtain a set of
class-agnostic regions $\{s_i\}$ that over-segment the scene. The composite is a phenology-aware
recoloring: a per-pixel first-harmonic NDVI fit mapped to HSV color (hue from seasonal phase,
saturation from amplitude, value from mean NDVI), so phenologically distinct fields take on
separable colors. The region boundaries are trustworthy, but a single parcel is usually split
into several regions. This deliberately trades
recall for boundary precision, and defers the ``how many parcels'' decision to the crop-aware
merge in (C) rather than to SAM.

\paragraph{(B) Harmonic phenological signature.}
For each region $s_i$ we form two temporal series, the optical NDVI~\citep{tucker1979ndvi} and
the SAR backscatter, each averaged over the region's pixels at every acquisition. We fit a
low-order (double) harmonic model to each series,
\[
f(\tau)=a_0+\sum_{h=1}^{2}\bigl[a_h\cos(2\pi h\tau/P)+b_h\sin(2\pi h\tau/P)\bigr],
\]
with $P$ the season length. From both NDVI and SAR we read off the interpretable coefficients:
the seasonal mean, the first- and second-harmonic amplitudes, and their phases. Their
concatenation is the region's phenological signature $\phi(s_i)$. The fit is analytic least
squares, robust to irregular revisit and missing dates, and needs no training. Its amplitudes and
phases encode \emph{when} a region greens up and dries down, which is what separates crops.
Fitting on NDVI\,+\,SAR rather than raw reflectance keeps the signature calibration-independent,
and the ablation shows that radar is what most lifts typing accuracy.

\paragraph{(C) Graph-energy merge into parcels.}
To recover parcels from over-segmentation, we build an adjacency graph $G=(V,E)$ over regions
and seek a labeling $\ell$ that groups phenologically similar neighbors while respecting SAM
boundaries. We minimize a Potts energy
\[
E(\ell)=\sum_{i} D_i(\ell_i)+\lambda\!\!\sum_{(i,j)\in E}\! w_{ij}\,[\ell_i\neq \ell_j],
\]
with edge weight $w_{ij}=\exp(-\|\phi(s_i)-\phi(s_j)\|^2/\sigma^2)$. We solve it with standard
graph-cut moves~\citep{boykov2001graphcuts}, and the connected components of the resulting
labeling are the merged parcels. The pairwise term fuses regions with close signatures, while the
boundary weight $w_{ij}$ makes a merge cheap across a weak, phenologically consistent seam and
expensive across a strong one. In the ablation, this energy-based merge is both more accurate and
less fragmented than an appearance-only IoU-NMS merge at identical inputs.

\paragraph{(D) Few-shot prototype typing.}
Only now do labels enter, and only $k$ per class. For class $c$ we average the signatures of its
$k$ labeled parcels into a prototype $p_c$. Each merged parcel $P$ is then typed by its nearest
prototype under a standardized Euclidean metric, $\hat c(P)=\arg\min_{c}\|\phi(P)-p_c\|$, with
features $z$-scored so that each signature dimension contributes comparably. Because $\phi$ is a
compact phenological descriptor, a handful of exemplars suffices. In the label-free ($k{=}0$)
limit we replace prototypes by an agronomic calendar and switch to cosine similarity over an
informative subset of the signature (the seasonal mean and first-harmonic phase), which already
exceeds an open-vocabulary zero-shot segmenter.

\paragraph{(E) Topology closure.}
A final closure step fills small gaps left by over-segmentation, removes slivers, and enforces
parcel connectivity, producing a clean panoptic map in which each field is a single typed instance.

\paragraph{Why training-free.}
Every stage is either a frozen model (A), an analytic fit (B), a combinatorial optimization
(C, E), or nearest-prototype matching from $k$ labels (D). No gradients are computed on crop
labels, so PhenoStitch operates directly in the label-scarce regime and applies to a new domain
without any architecture retraining. Only a few labeled parcels per class in that domain are
needed to build prototypes. Algorithm~\ref{alg:phenostitch} summarizes the full procedure.

\begin{algorithm}[t]
\caption{PhenoStitch (training-free, one patch)}
\label{alg:phenostitch}
\textbf{Input}: SITS $X=\{(I_t,\tau_t)\}_{t=1}^{T}$; class prototypes $\{p_c\}_{c=1}^{C}$ built
once from $k$ labeled parcels/class.\\
\textbf{Output}: panoptic map (a crop class and a parcel instance per pixel).
\begin{algorithmic}[1]
\STATE $\{s_i\} \gets \textsc{SAM}(\textsc{Composite}(X))$ \COMMENT{(A) label-free over-seg.}
\FOR{each region $s_i$}
  \STATE $\phi(s_i) \gets \textsc{HarmonicFit}(\mathrm{NDVI}(s_i),\,\mathrm{SAR}(s_i))$ \COMMENT{(B) signature}
\ENDFOR
\STATE build region graph $G$ with weights $w_{ij}\!=\!e^{-\|\phi(s_i)-\phi(s_j)\|^2/\sigma^2}$
\STATE $\ell \gets \arg\min_\ell E(\ell)$ by graph cuts \COMMENT{(C) graph-energy merge}
\STATE $\mathcal{P} \gets$ connected components of $\ell$ \COMMENT{parcels}
\FOR{each parcel $P \in \mathcal{P}$}
  \STATE $\hat c(P) \gets \arg\min_c \|\phi(P)-p_c\|$ \COMMENT{(D) few-shot typing}
\ENDFOR
\STATE \textbf{return} $\textsc{TopologyClosure}(\mathcal{P},\hat c)$ \COMMENT{(E) clean panoptic map}
\end{algorithmic}
\end{algorithm}

\paragraph{Complexity and cost.}
PhenoStitch has no training phase and no learned weights beyond the frozen SAM backbone. Per
patch, the cost is a single SAM forward pass, one closed-form harmonic fit per region (linear
least squares over the available acquisitions), one graph cut on the region-adjacency graph, and
a nearest-prototype lookup per parcel; each step is linear or near-linear in the number of
regions and pixels. Prototypes are built once from the $k$ labels and reused, so adapting to a new
region or year costs only prototype construction, not retraining or backpropagation.

\section{Experiments}

\begin{figure*}[t]
\centering
\includegraphics[width=\textwidth]{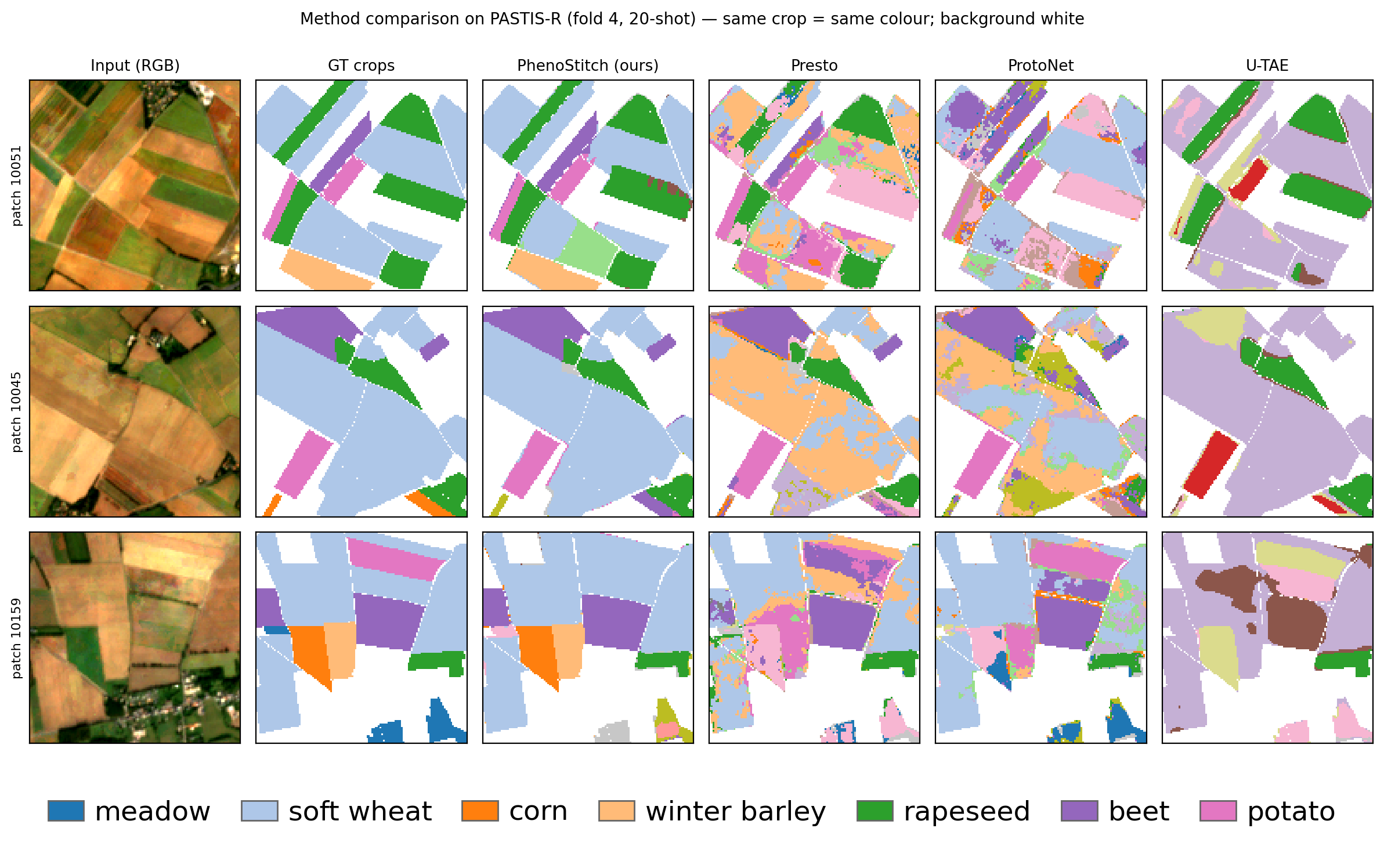}
\caption{\textbf{Qualitative comparison on PASTIS-R} (fold 4, $k{=}20$; same crop $=$ same color,
white background). Columns: input RGB, ground truth, PhenoStitch (ours), and the strongest
baselines. PhenoStitch recovers clean, correctly typed parcels matching the ground truth; the
per-pixel baselines (Presto, ProtoNet) are fragmented and mis-type whole fields, and the
matched-budget U-TAE collapses onto a few dominant classes.}
\label{fig:qual}
\end{figure*}

\begin{figure*}[t]
\centering
\begin{minipage}[b]{0.44\textwidth}
\centering
\includegraphics[width=\linewidth]{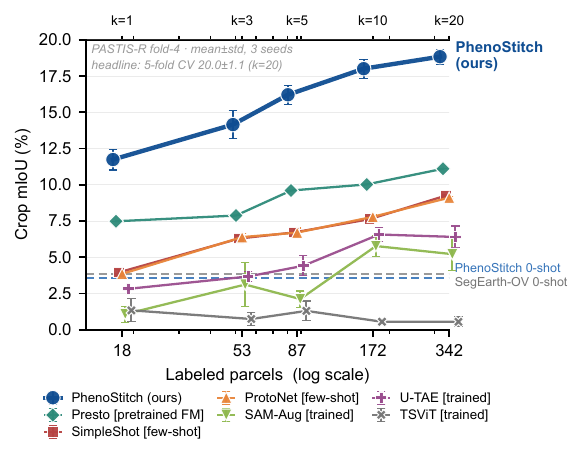}\\
{\small\textbf{(a)} crop mIoU vs.\ label budget}
\end{minipage}\hfill
\begin{minipage}[b]{0.50\textwidth}
\centering
\includegraphics[width=0.86\linewidth]{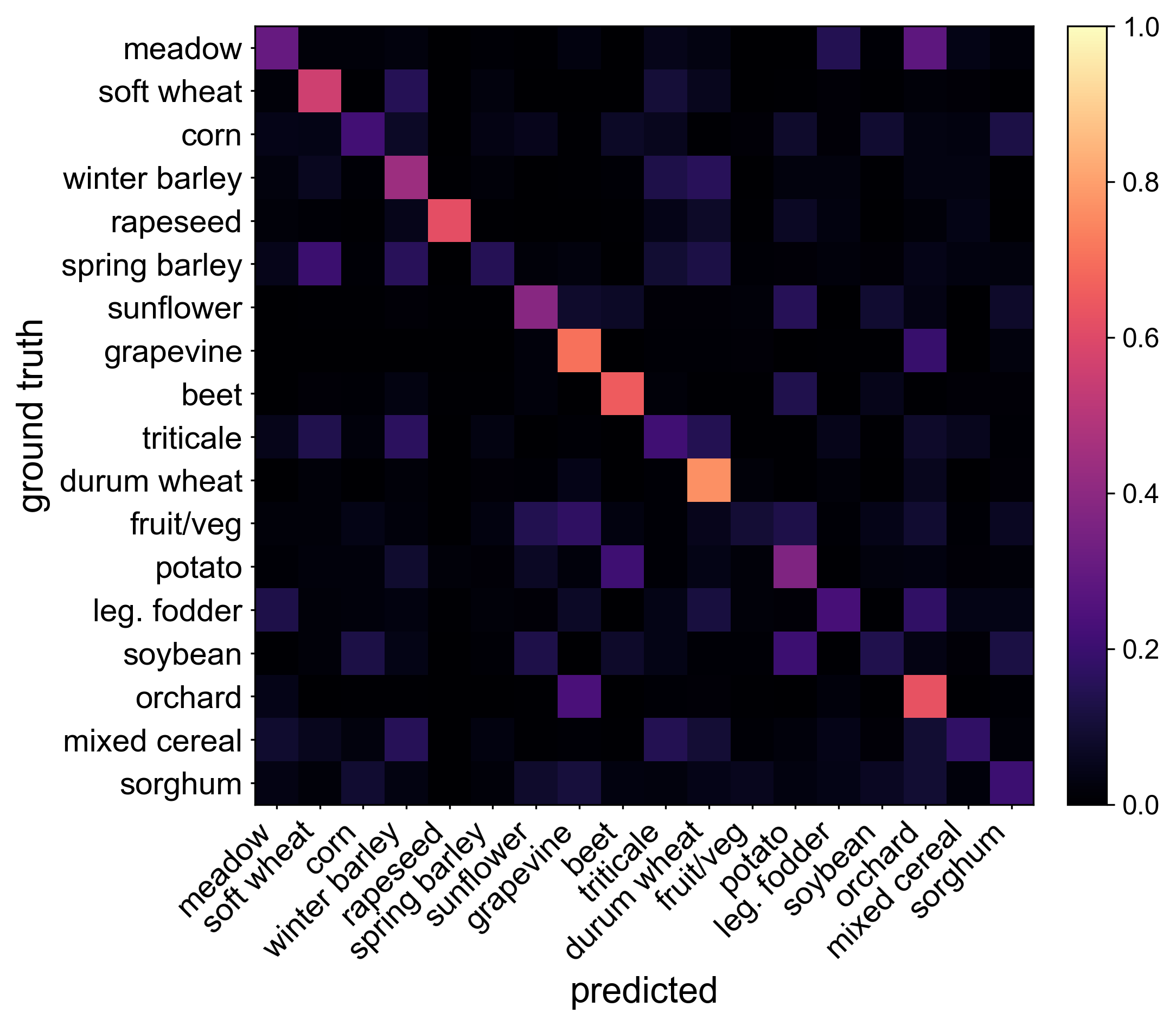}\\
{\small\textbf{(b)} PhenoStitch confusion matrix ($k{=}20$)}
\end{minipage}
\caption{\textbf{Label-scarce results and error analysis.}
\textbf{(a)} PhenoStitch (blue) leads at every budget $k\in\{1,3,5,10,20\}$ (full fold-4,
mean$\pm$std over $3$ seeds; the headline is the $5$-fold CV $20.0\!\pm\!1.1$ at $k{=}20$);
dashed lines are zero-shot references.
\textbf{(b)} Row-normalized confusion: phenologically distinct crops (rapeseed, beet, soybean,
corn, potato, orchard) are typed almost perfectly, while residual error concentrates among
near-identical winter cereals (soft wheat, winter barley, triticale, mixed cereal), which sets
the fine-grained ceiling.}
\label{fig:results}
\end{figure*}

\paragraph{Datasets.}
\textbf{PASTIS-R}~\citep{garnot2021panoptic} provides $128{\times}128$ Sentinel-2\,+\,Sentinel-1
SITS with panoptic (parcel + $18$-class) annotations over France. \textbf{ZueriCrop}
~\citep{turkoglu2021zuericrop} provides Sentinel-2 SITS over Switzerland with a $4$-level crop
hierarchy; we use it as an independent second benchmark (S2-only, finer taxonomy).

\paragraph{Protocol.}
We evaluate in the label-scarce regime: $k$ labeled parcels per class and no model training.
Our headline is $k{=}20$ ($\approx\!342$ parcels, $<\!1\%$ of labels) on PASTIS-R under
\textbf{$5$-fold $\times\,3$-seed cross-validation} (15 runs; each test fold uses a disjoint
support fold, over $3$ support-sampling seeds); we report mean\,$\pm$\,std. Delineation uses
zero labels throughout.

\paragraph{Metrics.}
We report crop \textbf{mIoU} over the fine ($18$-class) taxonomy and a coarse (agronomic
family) re-aggregation, plus the panoptic triplet \textbf{SQ} (parcel-delineation quality),
\textbf{RQ} (recognition), and \textbf{PQ}$=$SQ$\times$RQ~\citep{kirillov2019panoptic}. For
methods that output only a semantic map, instances are taken as connected components per class
(a PaPs~\citep{garnot2021panoptic} stand-in), so \emph{every} method, ours and all baselines,
receives a well-defined PQ under identical instancing. Coarse mIoU is re-aggregated
from the saved confusion matrix, not a re-run.

\paragraph{Baselines.}
Matched to the same $k$-label budget: a frozen foundation model
(Presto~\citep{tseng2023presto}); prototypical~\citep{snell2017protonet},
SimpleShot~\citep{wang2019simpleshot}, and transductive
$\alpha$-TIM~\citep{veilleux2021alphatim} few-shot classifiers; official
U-TAE~\citep{garnot2021panoptic} and TSViT~\citep{tarasiou2023tsvit} trained on the same tiny
budget; SAM-augmented training~\citep{samaug2026}; and zero-shot references
(PhenoStitch-0shot; SegEarth-OV~\citep{li2025segearth}). Trained baselines use the official
architectures with standard augmentation and early stopping at the matched $k$-label budget;
their low scores reflect the difficulty of fitting high-capacity temporal models to
$\sim\!1\%$ of labels, not a handicapped configuration.

\paragraph{Implementation details.}
SAM uses a frozen ViT-B backbone. Each region's NDVI and Sentinel-1 series are fit with two
harmonics by least squares; the graph-energy merge uses smoothness weight $\lambda{=}4$ and a
signature bandwidth $\sigma$ tuned once on a held-out patch; few-shot typing uses a standardized
Euclidean nearest-prototype rule, while the zero-shot calendar variant uses cosine similarity on
an informative subset of signature dimensions (seasonal mean and first-harmonic phase). Topology
closure fills gaps below a small area
threshold and assigns orphan pixels to the nearest parcel. All hyperparameters are shared
across folds, seeds, and both datasets, and were fixed once on a single held-out patch by
held-out crop mIoU; Table~\ref{tab:abl} reports the ranges swept for the main choices
(graph-energy weight $\lambda\in\{1.5,4,8\}$, harmonic order $1$--$3$, Euclidean versus cosine
typing, and the signature band set). Code and configurations will be released. All experiments
run on a single NVIDIA RTX~3090 ($24$\,GB) under Windows~11, with Python~3.12, PyTorch~2.2
(CUDA~12.1), NumPy~1.26, SciPy~1.17, and scikit-image~0.22: PhenoStitch performs no gradient
updates and needs only the frozen SAM forward pass, while the trained baselines are optimized
on the same hardware under the matched label budget.

\begin{table}[t]
\centering
\small
\setlength{\tabcolsep}{4pt}
\begin{tabular}{lccccc}
\toprule
Method & Sup. & mIoU & coarse & SQ & PQ \\
\midrule
\textbf{PhenoStitch (ours)} & free & \textbf{20.0}\,\small{$\pm$1.1} & \textbf{31.5} & \textbf{76.2} & \textbf{6.2} \\
Presto & FM & 11.7\,\small{$\pm$1.7} & 26.5 & 67.8 & 0.4 \\
SimpleShot & FS & 9.1\,\small{$\pm$1.8} & 19.3 & 65.0 & 0.3 \\
ProtoNet & FS & 9.0\,\small{$\pm$1.8} & 19.3 & 64.7 & 0.3 \\
$\alpha$-TIM & FS & 6.7\,\small{$\pm$1.1} & 15.3 & 63.8 & 0.2 \\
U-TAE & train & 5.3\,\small{$\pm$1.6} & 15.3 & 65.0 & 0.4 \\
SAM-Aug & train & 4.5\,\small{$\pm$1.5} & 15.5 & 66.5 & 0.3 \\
TSViT & train & 0.5\,\small{$\pm$0.3} & 4.3 & $\sim$4 & 0.0 \\
\midrule
PhenoStitch-0shot & zero & 3.8\,\small{$\pm$0.2} & 10.3 & 70.9 & 1.7 \\
SegEarth-OV & zero & 3.9 & -- & -- & -- \\
\bottomrule
\end{tabular}
\caption{\textbf{PASTIS-R, $5$-fold$\times3$-seed CV, $k{=}20$ ($<\!1\%$ labels).}
Crop mIoU (fine 18-class) / coarse (family) / SQ / PQ. Sup.: free=training-free,
FM=frozen foundation model, FS=few-shot, train=trained on the matched budget.
Full-supervision literature anchors (100\% labels): U-TAE $63.1$, TSViT $65.4$ mIoU, and the
purpose-built panoptic U-TAE+PaPs at $\mathrm{PQ}=40.4$.}
\label{tab:main}
\end{table}

\subsection{Main result: SOTA in the label-scarce regime}
Table~\ref{tab:main} and Figure~\ref{fig:hero} report the headline comparison. PhenoStitch
reaches $20.0$ crop mIoU, $31.5$ coarse mIoU, $76.2$ SQ, and $6.2$ PQ, and is best on every
metric. It beats the frozen foundation model Presto ($11.7$ mIoU) by $\sim\!1.7\times$, the
few-shot classifiers ($9.0$--$9.1$) by $\sim\!2.2\times$, and the matched-budget trained
networks (U-TAE $5.3$, SAM-Aug $4.5$, TSViT $0.5$) by $\sim\!4$--$40\times$. The margin is not a
sampling artifact: PhenoStitch wins on \emph{all $15$} CV runs against \emph{every} baseline,
and against the strongest baseline, Presto, the paired difference is $8.3$ mIoU (Wilcoxon
signed-rank $p<10^{-4}$; paired $t$-test $p<10^{-10}$).

The panoptic gap is starker still. Because the semantic baselines emit no instances, their
$\mathrm{RQ}\!\approx\!0$ and thus $\mathrm{PQ}\!\approx\!0$ regardless of mIoU. This is not a
definitional artifact. Every method is instanced by the same per-class connected components, yet
the baselines' noisy maps still give $\mathrm{PQ}\!\le\!0.4$, because their regions rarely match
true parcels at $\mathrm{IoU}>0.5$. PhenoStitch's coherent parcels instead reach
$\mathrm{PQ}=6.2$. To confirm that connected components do not handicap the baselines, we also
trained a \emph{learned} parcel detector at the same $k{=}20$ budget: a U-TAE with an auxiliary
parcel-boundary head, instanced by seeded watershed (a lightweight PaPs~\citep{garnot2021panoptic}).
It reaches only $\mathrm{PQ}=0.5$, about $12\times$ below PhenoStitch and indistinguishable to two
decimals from connected-components instancing on the same backbone. At $\sim\!1\%$ labels the
boundary head learns almost no boundaries and the watershed degenerates to connected components,
so the panoptic gap does not depend on how baselines are instanced.

Qualitatively (Figure~\ref{fig:qual}), PhenoStitch produces clean, parcel-coherent maps that
track the ground truth, whereas the per-pixel baselines are fragmented and mis-type entire
fields. Delineation is strong and stable across folds ($\mathrm{SQ}=76.2\pm0.4$), confirming that
the label-free boundary leg is not the bottleneck.

\paragraph{Family-level accuracy.}
Grouping the $18$ classes into $7$ agronomic families roughly doubles every method's mIoU
(Table~\ref{tab:main}, coarse column). PhenoStitch reaches $31.5$ coarse mIoU, still well ahead
of Presto ($26.5$) and the few-shot classifiers (about $19$). The near-$2\times$ fine-to-coarse
gap, shared by all methods, indicates that most fine-grained errors are confusions \emph{within}
a family, such as among winter cereals, rather than gross mistakes across unrelated crops. For
applications that need crop groups rather than exact species, PhenoStitch already yields usable
maps at $\sim\!1\%$ labels.

\subsection{Label-budget curve}
Figure~\ref{fig:results}(a) sweeps the budget $k\in\{1,3,5,10,20\}$ ($3$ seeds each). PhenoStitch
dominates at every budget, from $11.7$ mIoU at $k{=}1$ (18 parcels) to $18.8$ at $k{=}20$ on
the single fold, and its error bars are tight. Even at $k{=}1$ it exceeds every baseline's
$k{=}20$ accuracy, and its zero-shot variant already surpasses the SegEarth-OV open-vocabulary
reference. The trained networks stay near or below the zero-shot line across the whole scarce
range, i.e.\ the matched-budget supervised regime never leaves the floor.

\subsection{Second dataset: ZueriCrop}
Table~\ref{tab:zueri} repeats the comparison on ZueriCrop ($k{=}5$, S2-only, 76 fine classes;
$500$ evaluation tiles). The absolute numbers are lower (finer taxonomy, no radar), but the
ranking is identical to PASTIS-R and PhenoStitch again leads on every metric, with
$\mathrm{SQ}=76.2$, as high as on PASTIS-R, showing the delineation leg is not dataset-specific. Consistent behavior on a second, independently collected dataset (a different
country and taxonomy) is what a training-free method should deliver, though we note this is
in-domain evaluation on each dataset rather than a direct cross-region transfer test.

\begin{table}[t]
\centering
\small
\setlength{\tabcolsep}{5pt}
\begin{tabular}{lccc}
\toprule
Method & mIoU (76) & coarse (16) & SQ \\
\midrule
\textbf{PhenoStitch} & \textbf{2.68} & \textbf{10.82} & \textbf{76.2} \\
Presto & 1.83 & 6.20 & 67.4 \\
ProtoNet & 1.27 & 5.96 & 65.3 \\
SimpleShot & 1.13 & 5.06 & 65.7 \\
$\alpha$-TIM & 0.53 & 3.17 & 66.5 \\
U-TAE & 0.16 & 1.66 & $\sim$0 \\
SAM-Aug & 0.05 & 1.32 & $\sim$0 \\
TSViT & 0.00 & 0.00 & $\sim$0 \\
\bottomrule
\end{tabular}
\caption{\textbf{ZueriCrop}, $k{=}5$, S2-only. Same ranking as PASTIS-R; delineation
(SQ) transfers unchanged.}
\label{tab:zueri}
\end{table}

\subsection{Ablations}
Table~\ref{tab:abl} ablates every design choice at full fold-4, $k{=}20$, reporting the change
from our full pipeline. \textbf{Radar matters most}: removing Sentinel-1 and typing on optical
NDVI alone costs $9.3$ fine mIoU while leaving SQ essentially unchanged ($-0.7$), so SAR drives
\emph{typing}, not delineation, and the two legs are decoupled. \textbf{Keeping the signature
compact} is next: adding red-edge and SWIR harmonics costs $6.9$ mIoU and a third harmonic $1.6$,
both curse-of-dimensionality effects in the few-shot regime; a single harmonic is statistically
tied ($+0.2$), and we keep the double harmonic because it also captures double-cropping. The
\textbf{graph-energy merge} beats an appearance-only IoU-NMS merge by $1.5$ mIoU at identical
prototypes and proposals, with lower fragmentation. The standardized \textbf{Euclidean}
nearest-prototype rule beats cosine by $1.9$, which is why the main results use it. The pipeline
is otherwise \textbf{robust}: the merge weight $\lambda$ is flat over $1.5$--$8$, and detrending
is within run-to-run noise.

\begin{table}[t]
\centering
\footnotesize
\setlength{\tabcolsep}{4pt}
\begin{tabular}{lccc}
\toprule
Configuration & mIoU & coarse & SQ \\
\midrule
\textbf{Ours} (full pipeline) & \textbf{13.6} & \textbf{23.1} & \textbf{75.1} \\
\midrule
\multicolumn{4}{l}{\quad\emph{change ($\Delta$) from ours:}} \\
\quad $-$ Sentinel-1 (S2-only) & $-9.3$ & $-12.4$ & $-0.7$ \\
\quad merge: IoU-NMS & $-1.5$ & $-1.5$ & $+0.1$ \\
\quad typing: cosine & $-1.9$ & $+0.7$ & $-2.2$ \\
\quad $+$ red-edge/SWIR bands & $-6.9$ & $-10.0$ & $+0.2$ \\
\quad harmonic order 1 & $+0.2$ & $-2.0$ & $+0.7$ \\
\quad harmonic order 3 & $-1.6$ & $-6.1$ & $+0.1$ \\
\quad merge $\lambda=1.5$ & $0.0$ & $0.0$ & $-0.2$ \\
\quad merge $\lambda=8$ & $-0.1$ & $0.0$ & $0.0$ \\
\quad no detrend & $+0.4$ & $+0.6$ & $-0.1$ \\
\bottomrule
\end{tabular}
\caption{\textbf{Ablations} (PASTIS-R, fold-4, $k{=}20$). The first row is our full pipeline
(S2$+$S1, energy merge, $2$ harmonics, standardized Euclidean typing, $\lambda{=}4$); every other
row reports the \emph{change} ($\Delta$) from it. Radar drives typing; a compact signature, the
graph-energy merge, and Euclidean typing all help, while $\lambda$ and detrending are robust.
For efficiency, ablations use a fixed $120$-patch support pool, so absolute numbers are below the
full-support headline (Table~\ref{tab:main}); the $\Delta$ comparisons hold within this setup.}
\label{tab:abl}
\end{table}

\subsection{Analysis and failure modes}
The confusion matrix (Figure~\ref{fig:results}b) shows that PhenoStitch's residual errors are
concentrated among phenologically near-identical winter-cereal sub-types (soft wheat, winter
barley, triticale, mixed cereal) and among rare crops with few pixels, not spread uniformly.
Phenologically distinct crops (rapeseed, beet, soybean, corn, potato) are typed almost perfectly.
Because these confusions stay within a family, the ceiling on fine accuracy is set by classes
that are genuinely hard to separate from seasonal signal alone, not by the pipeline: adding
red-edge/SWIR harmonics did not help and mildly hurt few-shot typing (curse of dimensionality).
Delineation, by contrast, stays near its ceiling on both datasets. Absolute accuracy in this regime is modest by design: $20$ mIoU is far below the
full-supervision ceiling near $65$. But in a target region without dense labels, the operative
alternative is not a supervised network; it is no map at all. Label-scarce accuracy is therefore
the quantity that matters there. These are the boundaries of the method: it delivers panoptic
maps and label-scarce SOTA, but not full-supervision accuracy on visually indistinct sub-types.

\section{Limitations and Future Work}
PhenoStitch has clear boundaries. First, its fine-grained ceiling is set by crops whose seasonal
signal is nearly identical, such as the winter cereals soft wheat, barley, and triticale, and by
rare classes with few pixels; separating these likely needs discriminative cues beyond a
low-order harmonic fit. Second, absolute accuracy in the label-scarce regime is modest, and the
method does not replace full supervision where dense labels already exist. Third, our evaluation
is in-domain on each dataset. Although nothing is trained on crop labels, a direct cross-region or
cross-year study, in which prototypes built in one country are applied to another, remains future work.
Fourth, delineation inherits the quality and biases of the frozen segmenter, and typing benefits
from radar, so accuracy may drop where Sentinel-1 is unavailable or where fields are too small for
the segmenter to resolve.

These boundaries point to concrete next steps. A label-efficient \emph{learned} signature or
merge could sharpen the winter-cereal distinctions without reintroducing dense supervision. A
formal cross-region transfer protocol would test the generalization that the training-free design
implies. Finally, extending the pipeline to smallholder and tropical systems, where labels are
scarcest and fields are smallest, is where a training-free method should matter most.

\section{Conclusion}
We recast panoptic crop mapping from a fully supervised task into a training-free, label-scarce
one. PhenoStitch decouples label-free parcel delineation (frozen SAM + graph-energy merge) from
few-shot crop typing (phenological signatures + prototypes). On $5$-fold$\times3$-seed
cross-validation over PASTIS-R and ZueriCrop, it is state-of-the-art on every panoptic metric at
$\sim\!1\%$ labels, beating foundation-model, few-shot, and matched-budget trained baselines. It
is also the only method to attain non-trivial panoptic quality. The evidence is bounded:
fine-grained accuracy is capped by phenologically near-identical crop sub-types rather than by
delineation, which stays stable at $76$ SQ across datasets. Because nothing is trained on crop
labels, PhenoStitch offers a practical route to panoptic mapping in the regions and years where
dense labels do not exist.

\bibliography{main}

\end{document}